\documentclass[sigconf]{acmart}
\AtBeginDocument{%
  \providecommand\BibTeX{{%
    \normalfont B\kern-0.5em{\scshape i\kern-0.25em b}\kern-0.8em\TeX}}}

\usepackage{multirow}
\usepackage{makecell}
\usepackage{balance}

\copyrightyear{2023} 
\acmYear{2023} 
\setcopyright{acmlicensed}\acmConference[MM '23]{Proceedings of the 31st ACM International Conference on Multimedia}{October 29-November 3, 2023}{Ottawa, ON, Canada}
\acmBooktitle{Proceedings of the 31st ACM International Conference on Multimedia (MM '23), October 29-November 3, 2023, Ottawa, ON, Canada}
\acmPrice{15.00}
\acmDOI{10.1145/3581783.3612508}
\acmISBN{979-8-4007-0108-5/23/10}

\begin{document}

\title{360-Degree Panorama Generation from Few Unregistered NFoV Images}

\author{Jionghao Wang}
\affiliation{%
\institution{Shanghai Jiao Tong University}
\country{}
}
\email{shanemankiw@sjtu.edu.cn}
\authornote{Indicates equal contribution.}

\author{Ziyu Chen}
\affiliation{%
\institution{Shanghai Jiao Tong University}
\country{}
}
\email{1252060456@sjtu.edu.cn}
\authornotemark[1]

\author{Jun Ling}
\affiliation{%
\institution{Shanghai Jiao Tong University}
\country{}
}
\email{lingjun@sjtu.edu.cn}

\author{Rong Xie}
\affiliation{%
\institution{Shanghai Jiao Tong University}
\country{}
}
\email{xierong@sjtu.edu.cn}

\author{Li Song}
\affiliation{%
\institution{Shanghai Jiao Tong University}
\country{}
}
\email{song_li@sjtu.edu.cn}
\authornote{Corresponding author. Affiliated with both School of Electronic Information and Electrical Engineering, Shanghai Jiao Tong University
\& MoE Key Lab of Artificial Intelligence, AI Institute, Shanghai Jiao Tong University.
\\
Jionghao Wang, Ziyu Chen, Jun Ling and Rong Xie are with School of Electronic Information and Electrical Engineering, Shanghai Jiao Tong University.}

\renewcommand{\shortauthors}{Jionghao Wang, Ziyu Chen, Jun Ling, Rong Xie, \& Li Song}
\begin{abstract}
360$^\circ$ panoramas are extensively utilized as environmental light sources in computer graphics. However, 
capturing a 360$^\circ$ $\times$ 180$^\circ$ panorama poses challenges due to the necessity of specialized and costly equipment, and additional human resources. Prior studies develop various learning-based generative methods to synthesize panoramas from a single Narrow Field-of-View (NFoV) image, but they are limited in alterable input patterns, generation quality, and controllability. To address these issues, we propose a novel pipeline called \textit{PanoDiff}, which efficiently generates complete 360$^\circ$ panoramas using one or more unregistered NFoV images captured from arbitrary angles. Our approach has two primary components to overcome the limitations. Firstly, a two-stage angle prediction module to handle various numbers of NFoV inputs. Secondly, a novel latent diffusion-based panorama generation model uses incomplete panorama and text prompts as control signals and utilizes several geometric augmentation schemes to ensure geometric properties in generated panoramas. Experiments show that PanoDiff achieves state-of-the-art panoramic generation quality and high controllability, making it suitable for applications such as content editing.
\end{abstract}

\ccsdesc[500]{Computing methodologies~Computer vision}
\ccsdesc[300]{Computing methodologies~Computer graphics}

\keywords{360-degree panorama, generative models, multimodal models, latent diffusion, image pose estimation}

\begin{teaserfigure}
    \centering
  \includegraphics[width=0.98\textwidth]{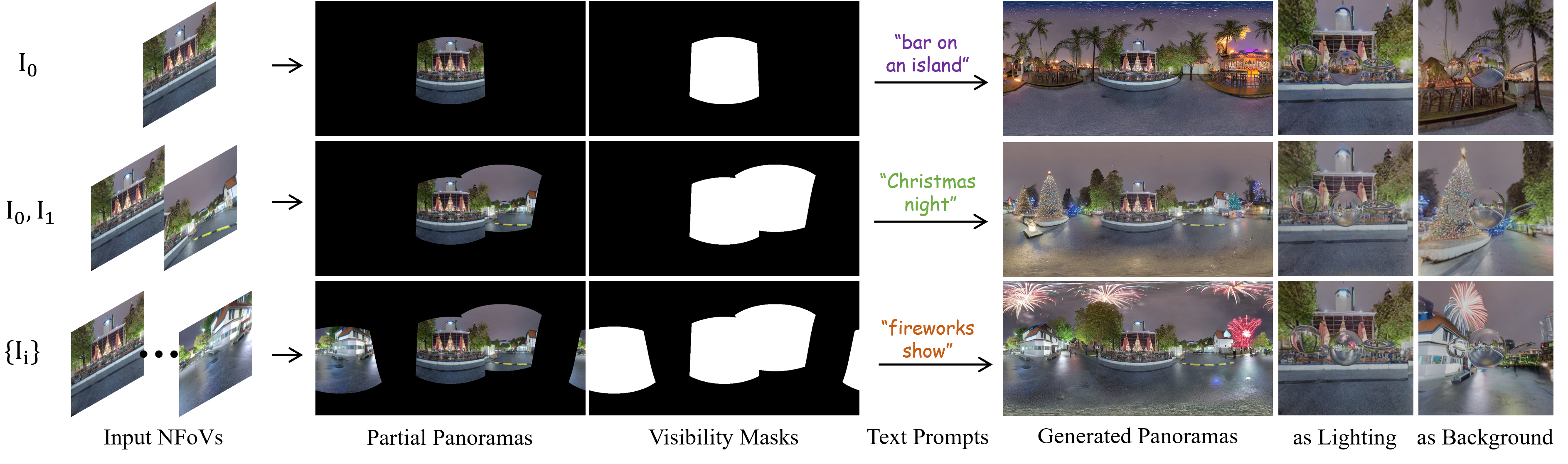}
  \caption{Illustration of our PanoDiff, a novel approach that is capable of synthesizing fine-grained and diverse 360-degree panorama from a(few) unregistered NFoV (Narrow Field of View) image(s) and text prompts.}
  \label{fig:teaser}
\end{teaserfigure}

\maketitle

\section{Introduction}

Panoramic images, which capture an extensive field of view encompassing a full $360^\circ$ horizontal by $180^\circ$ vertical FoV scene, have become increasingly significant across various applications, such as environment lighting, VR/AR, and autonomous driving system. However, obtaining high-quality panoramic images can be both time-consuming and costly, as they typically necessitate the use of specialized panoramic cameras or stitching software to combine images from multiple perspectives. 
Our method addresses two main limitations regarding previous generating methods, namely \textit{input pattern} and \textit{generation quality \& controllability}.

For \textit{Input pattern.} Most previous works~\cite{akimoto2022diverse,wang2022stylelight, dastjerdi2022guided} only support a single FoV region in the center of an incomplete panorama as input. However, relying solely on a single input region restricts flexibility when controlling specific contents of the generated scene. Such flexibility is particularly important for applications that require precise control over the visual elements and ensure accurate representations of the intended scene.

In terms of \textit{generation quality \& controllability.} Generating a complete 360-degree panorama from NFoV images could be viewed as a large-hole inpainting problem~\cite{suvorov2022resolution}. 
Previous methods~\cite{akimoto2022diverse,hara2022spherical,wang2022stylelight} typically all rely on GAN(Generative Adversarial Networks)~\cite{goodfellow2020generative} based methods. Besides, previous methods approach this problem as an image-conditioned generation task, leaving their pipeline with little control and flexibility over the generation results. 
A more recent work~\cite{dastjerdi2022guided} proposes to use additional text guidance for a GAN-based image inpainting method~\cite{zhao2021comodgan}.
However, diffusion-based image generation methods~\cite{ho2020denoising} have shown impressive results on various generative tasks, and models trained on large datasets~\cite{rombach2021highresolution, schuhmann2022laion} have shown better performance and robustness against GAN-based models~\cite{dhariwal2021diffusion, ho2022cascaded}. Moreover, GANs have limited mode coverage and are difficult to scale for modeling complex multimodal distributions. Unlike GANs, likelihood-based models such as diffusion models are capable of learning the complex distribution of natural images, resulting in the generation of high-quality images, as mentioned in ~\cite{dhariwal2021diffusion, rombach2021highresolution}. 

To overcome these limitations, a pipeline capable of accepting a flexible number of NFoV image(s) as input and generating high-fidelity panoramas is crucial. Nevertheless, this endeavor presents two main challenges: 1) estimating relative camera poses and accurately warping the input NFoV image(s) on the panorama, and 2) using a latent-diffusion-model-based method to generate the entire panorama from input partial panorama of various shapes.

This paper introduces \textit{PanoDiff}, a novel pipeline that efficiently generates complete 360° panoramas using one or more unregistered NFoV images captured from arbitrary angles. 
The proposed method overcomes the limitations of existing methods, it enables the generation of high-quality panoramas from incomplete 360-degree panoramas that warped from any number of NFoV inputs. 
This is achieved by addressing two key challenges. First, we propose a robust two-stage angle prediction pipeline that classifies image pairs based on their overlap before regressing specific angle values. Second, we train a hypernetwork that controls a pre-trained large latent-diffusion model, and utilizes geometric augmentation schemes during both training and inference sampling phases to ensure the geometric properties of the generated panoramas.

We summarize our contributions as follows. 
\textbf{Firstly}, the first flexible framework for generating panoramic images from single or multiple NFoV inputs. A two-stage pose estimation module is specially designed for relative pose estimation. 
\textbf{Secondly}, as far as we know, PanoDiff is the first latent-diffusion-based panorama outpainting model that can not only handle incomplete panoramas of various shapes as input but also supports text prompts.
\textbf{Finally}, PanoDiff outperforms existing methods in terms of quantitative and qualitative results on different input types, as demonstrated through abundant experiments.


\begin{figure*}[htbp]
  \centering
  \includegraphics[width=\linewidth]{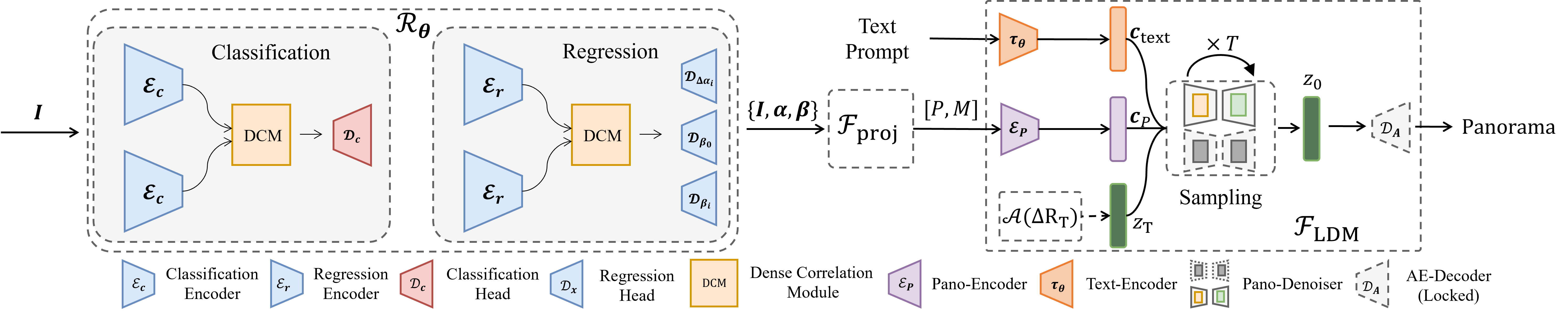}
  \caption{An overview of generating panorama from a few NFoV images. We first calculate their relative rotations based on a two-stage angle prediction network $\mathcal{R}_{\theta}$ (Sec. ~\ref{sec:poseesti}), then project them using backward equirectangular projection $\mathcal{F}_{\mathrm{proj}}$ (Sec. ~\ref{sec:panogen}) to obtain partial panorama $P$ and visibility mask $M$. Finally, we feed $[P, M]$ along with text prompts to our control-based latent diffusion model $\mathcal{F}_{\mathrm{LDM}}$ (Sec. ~\ref{sec:panogen}) and sample iteratively in our rotating schedule (Sec. ~\ref{sec:schedule}) to get the final generated panorama. }
  \label{fig:overview}
\end{figure*}
\section{Related Works}
\subsection{Rotation Estimation}
Camera pose estimation has been a long-standing task in computer vision~\cite{liao2023deep}. Most prior works focus on identifying various visual cues from pairs of images. For instance, the classic SIFT~\cite{lowe2004distinctive} extracts scale and direction invariant features from input images and matches them to find correspondences. Some approaches utilize neural networks for more robust feature extraction or graph neural networks for enhanced matching~\cite{detone2018superpoint, sarlin2020superglue, Rockwell2022}. However, these methods do not account for extreme rotations where the input images have little to no overlap, known as wide baseline scenarios. Some earlier works focus on mining specially handcrafted cues~\cite{ma2022virtual} to address this problem, but they are not generalizable for scenes without a specific clue. The work that inspired us the most is ~\cite{cai2021extreme}, which estimates pair-wise rotation angles in both overlap and wide baseline scenarios. However, the estimation network occasionally misidentifies an overlapping pair of images as non-overlapping, which severely impacts the generation quality later on.

\subsection{Diffusion Models}
Diffusion models have shown impressive ability in generative tasks~\cite{dhariwal2021diffusion, ho2020denoising}. Generally, diffusion models represent the image generative process as a denoising process, where a noise map sampled from white Gaussian noise is iteratively denoised using a learned prior distribution $p_{\theta}(x_{t-1} | x_{t})$. The objective function for training diffusion models is simplified as 
\begin{equation}
L_{\text {simple }}=E_{t, x_0, \epsilon}\left[\left|\epsilon-\epsilon_\theta\left(x_t, t\right)\right|^2\right]
\label{equ:simple}
\end{equation}
which has been shown to be approximately equivalent to optimizing the prior probability distribution's variational lower bound \cite{ho2020denoising}.

To further improve the efficiency and effectiveness of the diffusion model, recent work has proposed to conduct the denoising process in latent space instead of pixel space \cite{rombach2021highresolution}. In addition, they integrate multi-head cross-attention mechanism~\cite{vaswani2017attention, chen2021crossvit} in denoising U-Net~\cite{oktay2018attention, ronneberger2015u} blocks. This approach enables features of different modalities to guide the generation process\cite{ho2022classifier}, e.g. CLIP-ViT~\cite{Radford2021LearningTV}. Working on latent space along with some other sampling schedule~\cite{song2020denoising} could also potentially make the sampling more efficient. 
Controlling the pre-trained model parameters by training hypernetworks\cite{hyperguide,zhang2023adding} can improve its performance on a more specific task while preserving its original generative capability.

\subsection{Diffusion-based Inpainting}
The task of inpainting was predominantly been done by GAN-based models~\cite{liu2022partial, zhao2021large, suvorov2022resolution}. Recently, diffusion-based methods have been proposed and have already achieved promising results~\cite{lugmayr2022repaint, wang2022zero}. However, ~\cite{lugmayr2022repaint, wang2022zero} operate the diffusion and sampling process solely on image space rather than latent space, which limits their generation flexibility. Recently, ~\cite{runwayml} offered an image inpainting model finetuned from Stable Diffusion~\cite{rombach2021highresolution} and demonstrates strong text-controlled inpainting ability.

\subsection{Panorama Generation}
With recent advancements in deep generative neural networks, several works adopt different forms of GAN-based methods to generate full panoramas~\cite{dastjerdi2022guided, akimoto2022diverse, wang2022stylelight, liao2022cylin}, such as VQGAN~\cite{esser2021taming} and CoModGAN~\cite{zhao2021comodgan}. However, these methods are limited to single-view input, in contrast to our alterable input patterns. One work has utilized a few NFoV images as input~\cite{hara2022spherical} with accurate camera positions already given, thus cannot deal with unregistered camera inputs.

Recently, some researchers have utilized diffusion models to generate panoramas from text prompts alone~\cite{chen2022text2light}. However, this work does not support partial FoV as input, and thus users have limited control over the outcome.

\section{Relative Pose Estimation}
\label{sec:poseesti}
\subsection{Problem Formulation}
Unlike previous panorama generation pipelines, our method takes either single or multiple NFoV images as input. In the case of multiple NFoV (Narrow Field-of-View) images captured at the same location but in different poses, our objective is to estimate the relative poses between them. In the context of omnidirectional images, where no camera translation is involved, we formulate the relative camera poses as rotation angles. Specifically, we view the 360$^\circ$ panorama as a spherical \textit{map} and express the rotation angles as the 1) \textit{lookat} direction, which includes the longitude $\alpha$, latitude $\beta$, and 2) roll angle $\gamma$. For a single image, our pipeline estimates only $\beta$ and places it in lateral center, i.e., $\alpha=\gamma=0$. Formally, for a set of $N$ NFoV images $\boldsymbol{I} = {I_0, I_1, ..., I_{N-1}}$, we aim to learn a model $\mathcal{R}_{\theta}$ that predicts their relative angles:
\begin{equation}
\mathcal{R}_{\theta}(\boldsymbol{\mathrm{I}}) \to [\boldsymbol{\alpha}, \boldsymbol{\beta}, \boldsymbol{\gamma}]
\end{equation}
where $\boldsymbol{\alpha} = \{ \alpha_i\}_{i=0,...,N-1}$, $\boldsymbol{\beta} = \{ \beta_i\}_{i=0,...,N-1}$ and $\boldsymbol{\gamma} = \{ \gamma_i\}_{i=0,...,N-1}$, respectively.
We decompose this task into estimating pair-wise camera relative rotation, which is scalable and can be applied to pair input and sometimes more than two images. We design a pairwise angle prediction network that estimates the relative angles between images. For a set of $N$ NFoV images $(I_0, I_1, ..., I_{N-1})$, we select one image $I_0$ as an anchor and estimate the relative poses [$\Delta \boldsymbol{\alpha}$, $\Delta \boldsymbol{\beta}$, $\Delta \boldsymbol{\gamma}$] of the remaining images with respect to the anchor. Following the assumptions made in ~\cite{cai2021extreme}, we consider that cameras are typically upright, allowing us to estimate the absolute pitch angle (vertical/latitude) instead of the relative angle. We also assume that the camera roll angles are static and remain unchanged at $0$, i.e., $\forall i\in {0, ..., N-1}, \gamma_i = 0$. Consequently, our network $\mathcal{R}_{\theta}$ can be expressed as:
\begin{equation}
\mathcal{R}_{\theta}(I_0, I_i) \to [\Delta \alpha_{i \to 0}, \beta_{0}, \beta_{i}]
\end{equation}
where $\Delta \alpha_{i \to 0}$ is the relative longitude angle between the images, and $\beta_{0}, \beta_{i}$ are the absolute lattitude angle for $I_0$ and $I_i$, respectively.

\subsection{Two-stage Angle Prediction}
\label{sec:twostage}
\begin{figure}[htbp]
  \centering
  \includegraphics[width=\linewidth]{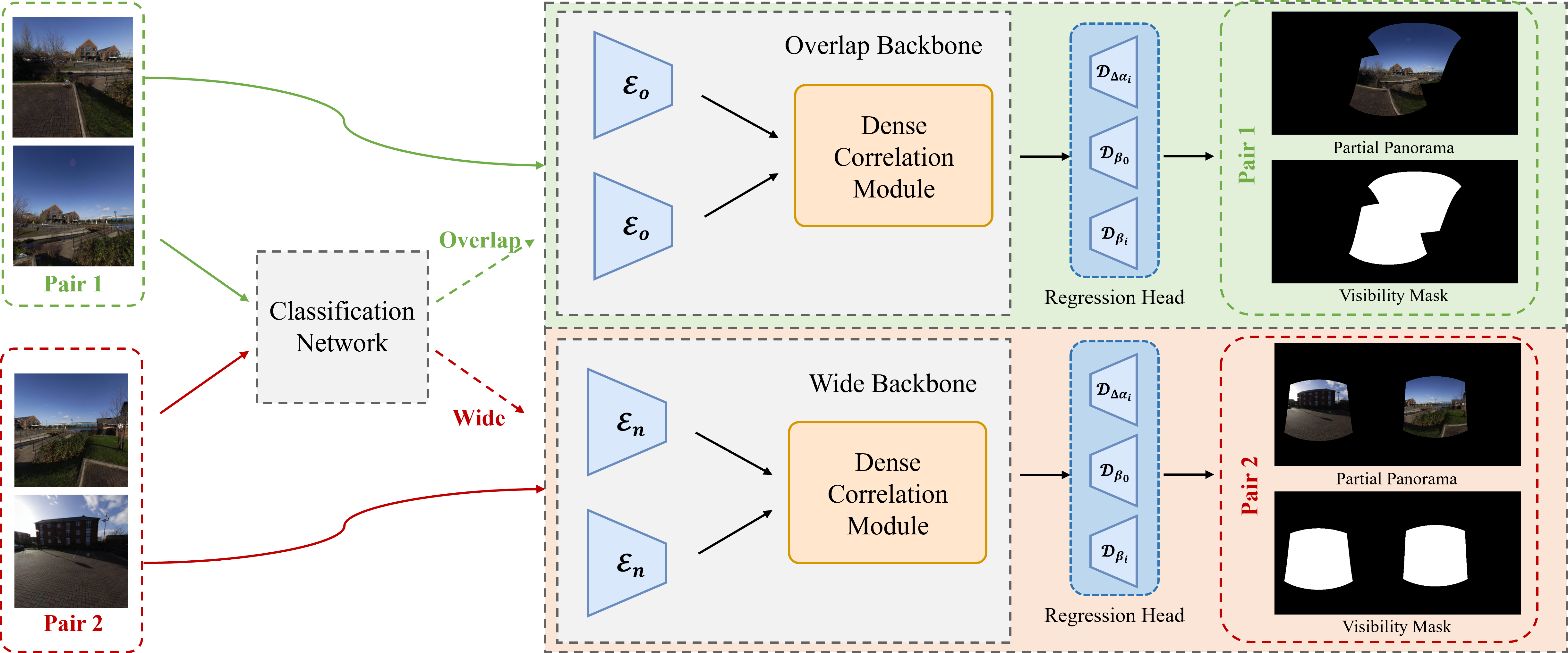}
  \caption{Illustration of our two-stage angle prediction network. ${\mathcal{E}_o, \mathcal{E}_n}$ represent feature encoders, and ${\mathcal{D}_{\Delta \alpha_i}, \mathcal{D}_{\beta_0}, \mathcal{D}_{\beta_i}}$ serve as prediction heads for overlap and wide-baseline angle regression, respectively. The pipeline first estimates whether the input image pair has an overlap, then directs the image pair to the corresponding regression network.}
  \label{fig:rotapipeline}
\end{figure}

In the context of FoV overlapping, we categorize image-pair relationships into two distinct types: \textit{overlap} and \textit{wide baseline}. In \textit{overlap} scenarios (green pair in Fig.~\ref{fig:rotapipeline}), two NFoV images share a portion of their FoV, resulting in an overlap on the panorama. Conversely, in \textit{wide baseline} situations (red pair in Fig.~\ref{fig:rotapipeline}), the NFoV image pair does not have overlapping FoV, leading to two separate regions on the panorama. 
The ultimate goal of our method is to create a plausible panorama for further applications. However, a single regression network~\cite{cai2021extreme} is not robust enough to differentiate between these two cases. Consequently, images without any overlap might sometimes be predicted to have an overlap, creating severe artifacts in the input for our controlled LDM, which would result in unsatisfactory results. Furthermore, the precision of a single regression network is insufficient.

To address these problems, we propose a two-stage angle prediction pipeline that initially performs classification to determine whether the two images have an overlap, and subsequently regresses the precise angles $[\Delta \alpha_{i \to 0}, \beta_{0}, \beta_{i}]$. This approach significantly reduces the issue of wide baseline images being mistakenly predicted to overlap. The overview of our rotation prediction module can be seen in Fig. ~\ref{fig:rotapipeline}.
We adopt the same backbone structure as ~\cite{cai2021extreme}, which consists of a feature encoder and a 4D dense correlation module. Within the two-stage pipeline, a total of three backbones are utilized, each with its own prediction head. To estimate the relative camera rotation between a pair of images, we first pass the image pair to our classifier to determine if the images are overlapping or wide apart. Subsequently, we feed the image pair to its corresponding regressor based on the classification result. 

In terms of training, our classifier is trained on our full set of training image pairs. A specialized penalty loss function is employed for misclassified cases, which effectively mitigates failure instances. As for our regressors, we separate our training image pairs into two splits based on their overlapping status and train the overlap branch and wide branch separately on their respective data splits. That is, for the overlap regressor, we train it using primarily overlap data pairs, and vice versa for the wide baseline regressor. 

\section{Panorama Generation}
\label{sec:panogen}

\subsection{Formulation}
\label{subsubsec:projection}

With Sec. ~\ref{sec:poseesti}, we can formulate a set of image-angle pairs, i.e., $\{\mathbf{\mathrm{I}_0}, 0, \beta_0\}$ and $\{\mathbf{\mathrm{I}_i}, \Delta \alpha_{i \to 0}, \beta_i\}_{i=1,\ldots,N-1}$. In this part, our goal is to take the images and their relative angles to produce a complete $360^{\circ} \times 180^{\circ}$ panorama. We divide this problem into two parts: 1) project the image set $\boldsymbol{I} = \{\mathbf{\mathrm{I}_i}\}$ onto the omnidirectional map based on their estimated angles $\boldsymbol{\alpha} = \{\alpha_i\}_{i=0,\ldots,N-1}$ and $\boldsymbol{\beta} = \{\beta_i\}_{i=0,\ldots,N-1}$; and 2) generate the full panorama using the incomplete input as a control signal. Specifically, we formulate this problem as:
\begin{equation}
\begin{aligned}
\mathcal{F}_{\mathrm{proj}}(\boldsymbol{I}, \boldsymbol{\alpha}, \boldsymbol{\beta}) &\to P, M, \\
\mathcal{F}_{\mathrm{LDM}}(P, M, \mathrm{text}) &\to \mathrm{Pano},
\end{aligned}
\end{equation}
where $\mathcal{F}_{\mathrm{proj}}$ and $\mathcal{F}_{\mathrm{LDM}}$ represent the projection and diffusion generation operations, respectively. The projection operation $\mathcal{F}_{\mathrm{proj}}$ takes the NFoV image set and their angles as input and produces a partial panorama image $P$ and a visibility mask $M$, indicating which parts of the FoV are missing. Subsequently, the LDM (Latent Diffusion Model) operation $\mathcal{F}_{\mathrm{LDM}}$ accepts the incomplete $P$, visibility mask $M$, and a text prompt as input, iteratively generating the full panorama $\mathrm{Pano}$.

\subsection{Recap: Controlling Stable Diffusion}
\label{sec:control}
We propose a generative process based on training a hypernetwork over a pre-trained Stable Diffusion (SD) model\cite{rombach2021highresolution}, a large text-to-image generative model utilizing latent diffusion. Latent diffusion models iteratively perform denoise sampling in latent space. At time step $t$, given input latent $z_t$ and condition $y$, the denoiser network is formulated as $\epsilon_\theta (z_t, t, y)$. The denoiser in SD employs a U-Net bottleneck architecture.

\cite{zhang2023adding} proposed to use combine trainable copies from pre-trained SD model parameters, and zero-convolution blocks, whose parameters are initialized as all zero. Concretely, suppose there is a neural network block $\mathcal{F}(\cdot;\Theta)$ from the pre-trained SD denoiser U-Net, which takes an input feature map $\mathbf{x} \in \mathrm{R}^{h \times w \times c}$ and outputs a feature $\mathbf{y}$, i.e. $\mathbf{y} = \mathcal{F}(\mathbf{x};\Theta)$. To add more conditions as control signals, two zero-convolution layers $\mathcal{Z}_1, \mathcal{Z}_2$ and their parameters $\Theta_{z1}, \Theta_{z2}$, and a trainable copy of the original neural block and its parameters as $\mathcal{F}_c$ and $\Theta_c$ are introduced. Now given a new condition signal $\mathbf{c}$, the new controlled output feature $\mathbf{y}_c$ can be written as:
\begin{equation}
\mathbf{y}_c = \mathcal{F}(\mathbf{x};\Theta) + \mathcal{Z}_2(\mathcal{F}_c(\mathbf{x} + \mathcal{Z}_1(\mathbf{c};\Theta_{z1});\Theta_c);\Theta_{z2}) 
\end{equation}

\begin{figure}[H]
  \centering
  \includegraphics[width=0.8\linewidth]{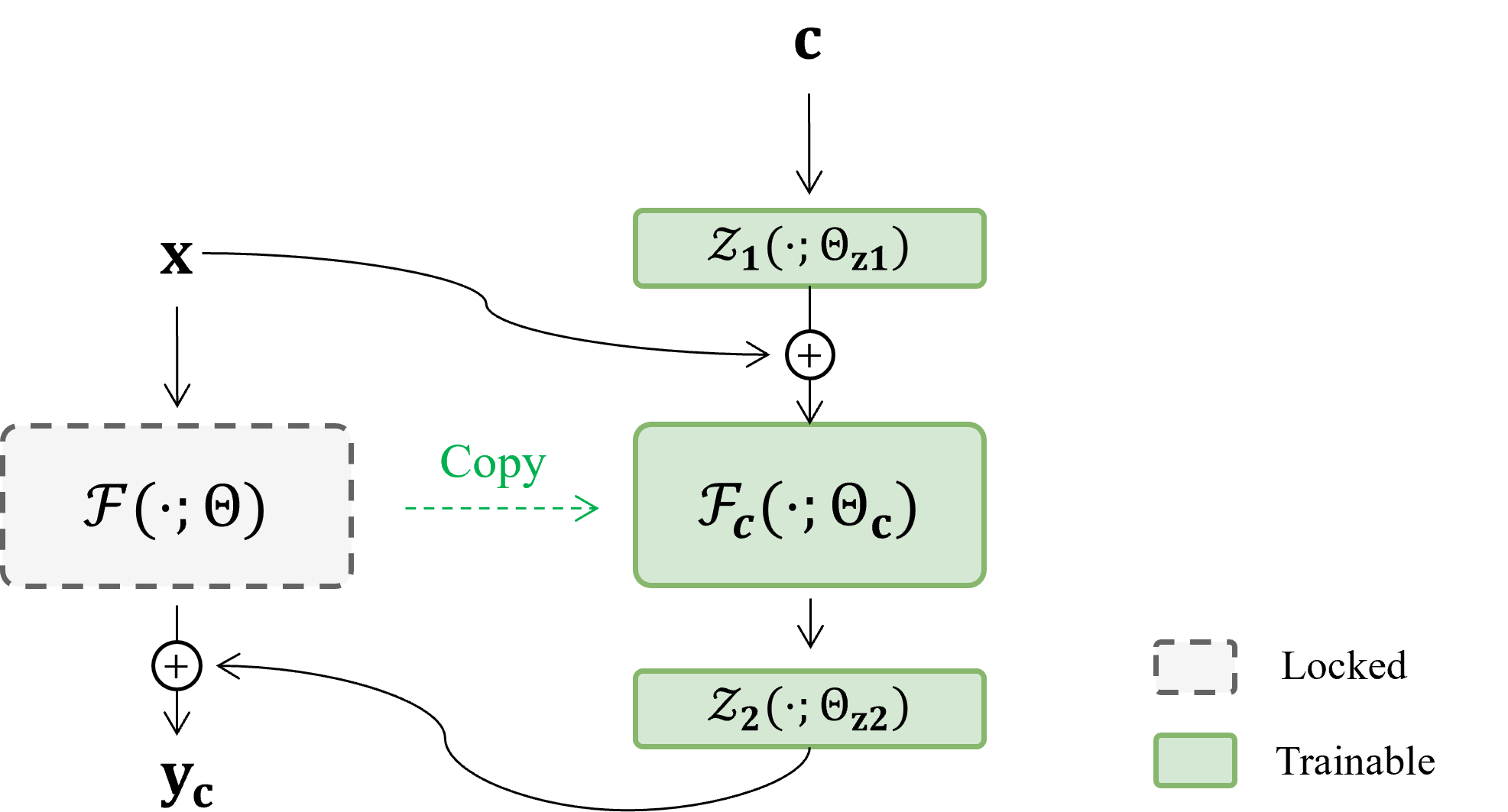}
  \caption{An intuitive explanation of integrating a new control signal into an existing model block.}
  \label{fig:control}
\end{figure}

It is worth noting that the original operation $\mathcal{F}(\cdot;\Theta)$ is locked and does not produce any gradient for backpropagation. In this setting, the new set of trainable parameters are ${\Theta_c, \Theta_{z1}, \Theta_{z2}}$, where the initial $\Theta_c$ is copied from $\Theta$, and both $\Theta_{z1}$ and $\Theta_{z2}$ are initialized as zeros. This allows the controlled input at the first step to be identical to the original output, i.e., $\mathbf{y}_c = \mathbf{y}$. This property is favorable as it maintains the generation ability of the pre-trained model and makes the updating process of ${\Theta_c, \Theta_{z1}, \Theta_{z2}}$ more stable. An intuitive illustration of this scheme can be seen in Fig.~\ref{fig:control}.

\subsection{Controlling LDM with Partial FoV}
Given a noisy latent space feature $z_t$ at time step $t$, we want to learn a denoiser that could predict the noise $\epsilon_t$ as in:
\begin{equation}
\epsilon_t = \epsilon_{\theta}(z_t, t, \mathbf{c}_{\mathrm{text}}, \mathbf{c}_{\mathrm{P}})
\end{equation}
Here, $\mathbf{c}_{\mathrm{text}}$ and $\mathbf{c}_{\mathrm{P}} = [M, P]$ represent the text conditions and input panorama conditions, respectively. We introduce two components for our partial-FoV-controlled LDM, namely Pano-Denoiser $\epsilon_\theta$ and Pano-Encoder $\mathcal{E}_P$.

Our Pano-Denoiser is designed as discussed in Sec. ~\ref{sec:control}, where we employ control units to introduce control signals to the pre-trained Stable Diffusion model. In practice, we follow the approach of \cite{zhang2023adding} and add control units to the four encoder blocks and one middle block of the denoising U-Net model used in Stable Diffusion, while utilizing zero-convolutions for the other four decoder blocks.

In addition to the Pano-Denoiser, we incorporate a shallow Pano-Encoder $\mathcal{E}_P$ to construct our controlling condition. In order to feed image-space signals, such as $P$ and $M$, into the latent space as control signals, we use $\mathcal{E}_p$ to encode them into latent features that are compatible with the original SD U-Net and consequently our control unit. The latent features are then passed to our learnable encoders and zero convolutions connected to the middle and decoder parts of the original SD U-Net.
\begin{figure}[htbp]
  \centering
  \includegraphics[width=\linewidth]{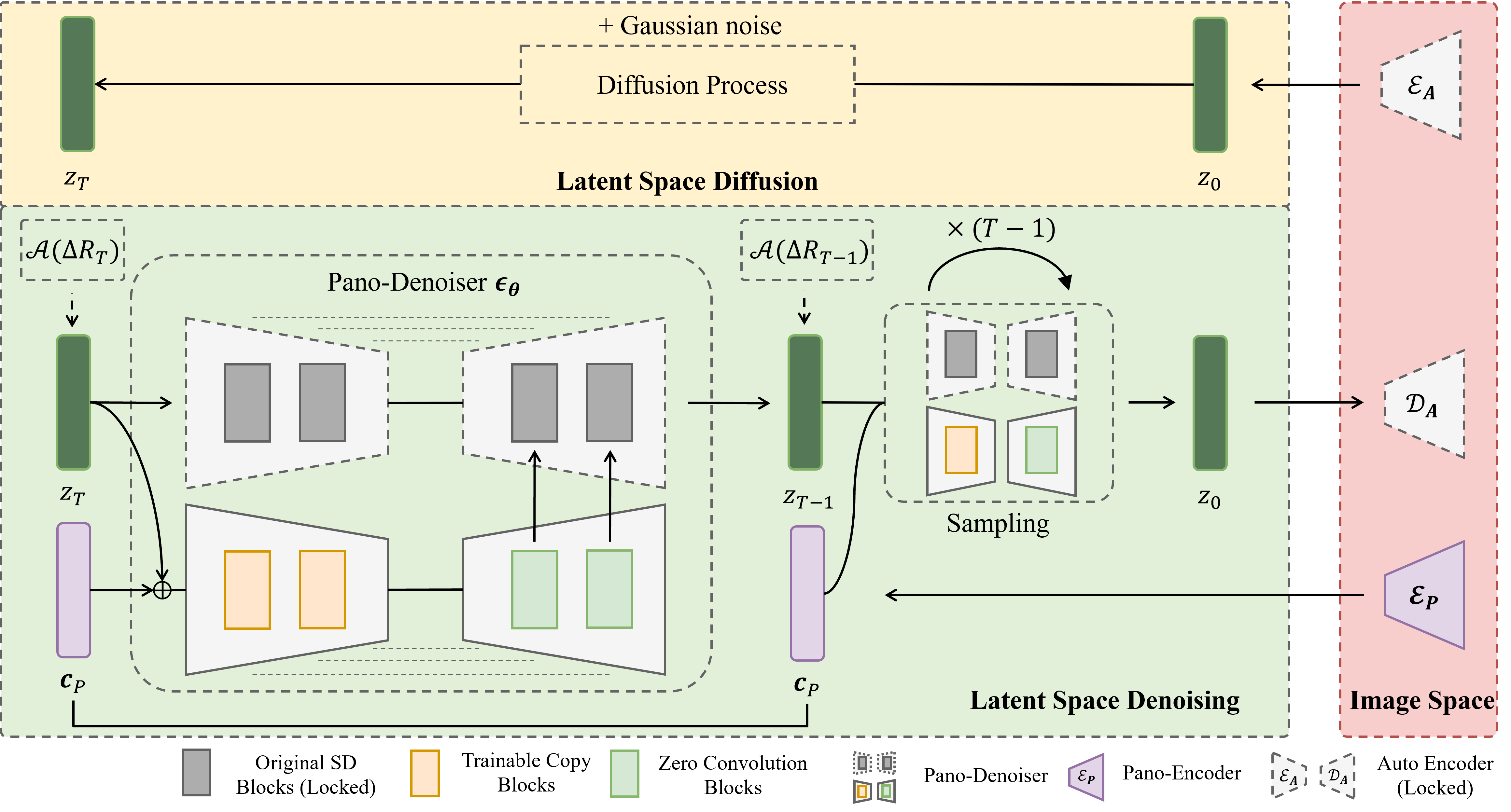}
  \caption{An overview of our partial-FoV-controlled latent diffusion model. The upper part and lower part are the diffusion process and sampling process of the model, respectively. Note that the text conditions and classifier-free guidance process are omitted from the figure for simplicity.}
  \label{fig:fullldm}
\end{figure}

As depicted in Fig. ~\ref{fig:fullldm}, during the diffusion process (upper part of the figure), the encoder $\mathcal{E}_A$ of the AutoEncoder transforms the image into feature space, and then Gaussian noises are added iteratively to the latent feature, as described in \cite{ho2020denoising}. For the denoising process, our Pano-Denoiser utilizes the partial panorama $P$ and visibility mask $M$ as control signals and denoises the input noisy latent $z_T$ iteratively in our rotating schedule(Sec. ~\ref{sec:schedule}) for $T$ steps until $z_0$ is generated. Finally, $z_0$ is decoded by the decoder $\mathcal{E}_D$ of the AutoEncoder to produce the final panorama.

\subsection{Denoising in 360-Degree}
In this subsection, we focus on denoising 360-degree panoramas while preserving their unique geometric characteristics. During the training stage, we introduce a rotation equivariance loss to enforce rotation consistency in the latent space. During the inference stage, we employ a customized rotating schedule to enhance the robustness and maintain the geometric integrity of the generated panoramas. Additionally, we implement a circular padding technique during inference to mitigate edge effects and prevent geometric discontinuity.
\subsubsection{Rotation Equivariance Loss}
\label{sec:equi}
Panoramas are captured to depict a completely spherical environment, which means they exhibit rotation equivariance. This property implies that if we apply a transformation based on an $SO(3)$ rotation (limited to 2 degrees of freedom corresponding to longitude and latitude rotations) to a panorama image, it should still be able to represent the same scene.

As our method functions within the latent space, the constraint on our denoiser can be expressed as:
\begin{equation}
\epsilon_{\theta}(\mathcal{A}(R) z_t, t, \mathbf{c}_{\mathrm{text}}, \mathcal{A}(R) \mathbf{c}_{\mathrm{P}}) = \mathcal{A}(R)\epsilon_{\theta}(z_t, t, \mathbf{c}_{\mathrm{text}}, \mathbf{c}_{\mathrm{P}}),
\label{equ:geo}
\end{equation}
where $\mathcal{A}(R)$ represents an image space projecting operation, similar to $\mathcal{F}_{\mathrm{proj}}$. This equation implies that when the latent feature $z_t$ and all input image-space conditions undergo a transformation by $\mathcal{A}(R)$, the predicted noise should also transform accordingly. To enforce this behavior, we introduce a rotation perturbation for latent features during the training phase. Consequently, the training objective evolves from Eq.~\ref{equ:simple} into:
\begin{gather*}
\mathcal{L} = \\
\mathbb{E}_{\boldsymbol{z}_0, t, \boldsymbol{c}_{\mathrm{text}}, \boldsymbol{c}_{\mathrm{P}}, \epsilon}[\| \mathcal{A}(\Delta R)\epsilon-\epsilon_\theta\left(\mathcal{A}(\Delta R)z_t, t, \boldsymbol{c}_{\mathrm{text}}, \mathcal{A}(\Delta R)\boldsymbol{c}_{\mathrm{P}}\right) \|_2^2]
\end{gather*}

\subsubsection{Rotating Schedule}
\label{sec:schedule}
\begin{figure}[!tbp]
  \centering
  \includegraphics[width=\linewidth]{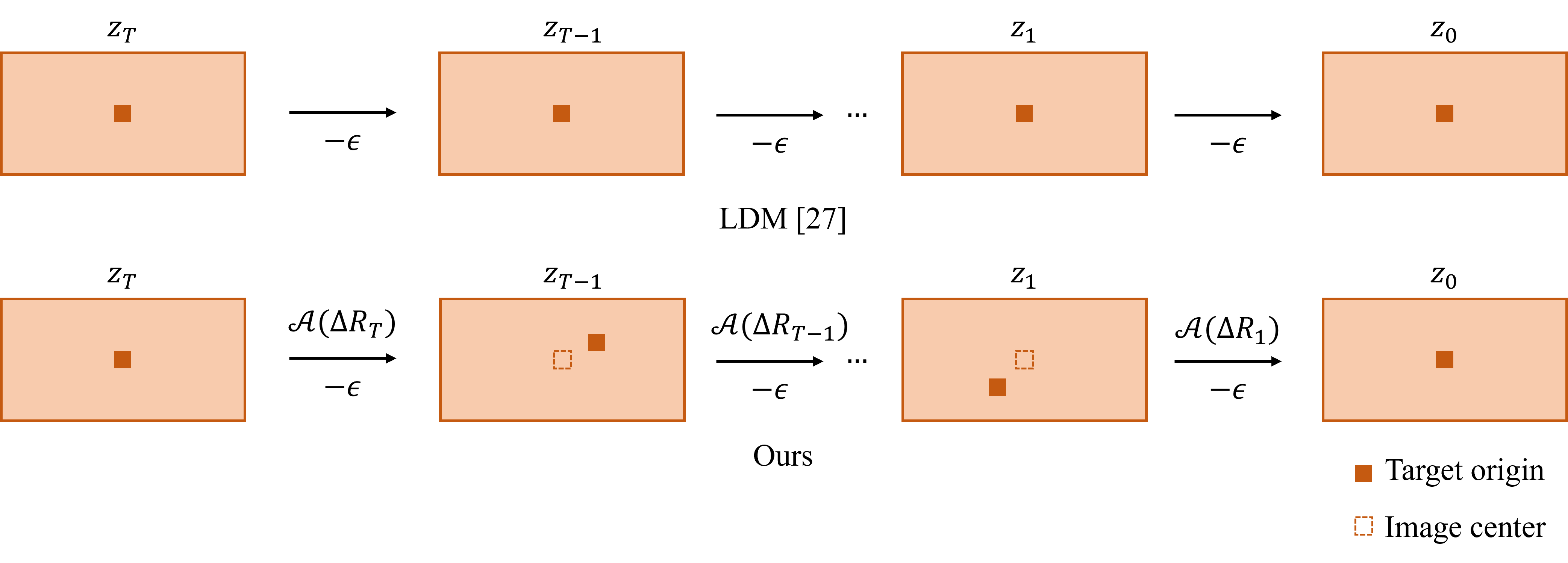}
  \caption{An illustration of our rolling schedule. The target origin represents the position of the original pixel corresponding to longitude and latitude coordinate $(\theta=0, \phi=0)$ on the panorama.}
  \label{fig:roll}
\end{figure}

During the inference process, we employ a customized schedule as depicted in Fig. ~\ref{fig:roll}. 

As illustrated in Figure~\ref{fig:roll}, the denoising steps are executed while the latent feature $z_i$ undergoes a scheduled transformation $\mathcal{A}(\Delta R_i)$ in a step-by-step manner. It is important to note that this schedule is consistent with the rotation constraint incorporated during the training phase, as both involve the same type of transformation in the latent space. The rotating schedule enhances the robustness of our method and facilitates the generation of panoramas with improved geometric integrity.

\subsubsection{Circular Padding}
\label{sec:cirpad}
\begin{figure}[htbp]
  \centering
  \includegraphics[width=0.8\linewidth]{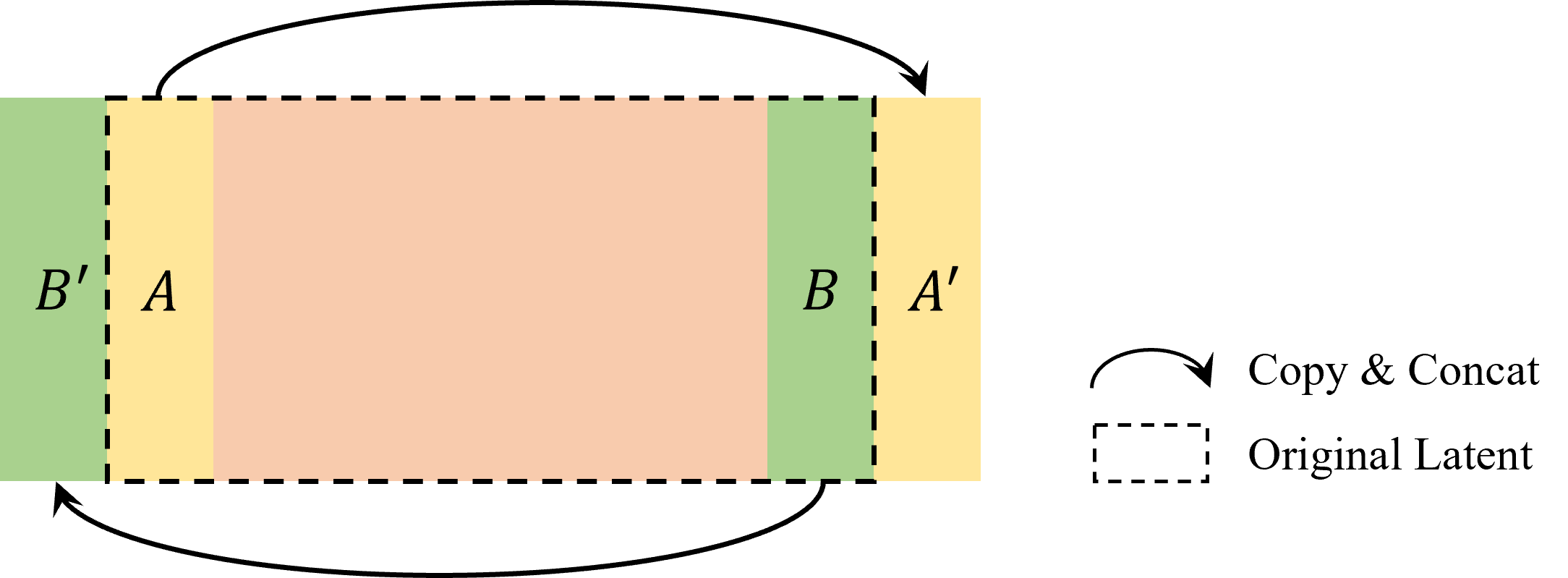}
  \caption{An illustration of our circular padding technique. Thin slices of the feature, denoted as $A$ and $B$, from both the left and right ends of the latent feature, are copied and concatenated to the right and left sides of the latent feature as $A^\prime, B^\prime$, respectively.}
  \label{fig:pad}
\end{figure}

As discussed in \cite{hara2022spherical}, generative methods operating in the latent space may lead to geometric discontinuity due to border effects of convolution operations. To address this issue, during inference time, we implement circular padding to mitigate edge effects. Specifically, the right portion of the latent feature is concatenated to the left side, while the left part of the original latent feature is concatenated to the right side. This process is illustrated in Fig. ~\ref{fig:pad}.

\begin{table*}[t]
\caption{FID$\downarrow$ results compared with other generation methods for quantitative evaluation.}
\label{tab:methodcompare}
\resizebox{0.73\linewidth}{!}{
\begin{tabular}{llccclccc}
\hline
                                &  & \multicolumn{3}{c}{SUN360~\cite{sun360}}                      &  & \multicolumn{3}{c}{Laval~\cite{lavalindoor}}                       \\ \cline{3-5} \cline{7-9}
Methods                          &  & Single        & Pair (GT rots) & Pair (Pred rots) &  & Single        & Pair (GT rots) & Pair (Pred rots) \\ \hline
SIG-SS~\cite{hara2022spherical}                          &  & 13.06         & 15.94         & 16.50           &  & -             & -             & -               \\
StyleLight~\cite{wang2022stylelight}                      &  & -             & -             & -               &  & 22.53             & 34.10             & 33.88               \\
Omni-Comp~\cite{akimoto2022diverse} &  & 14.69         & 12.38         & 12.63           &  & 18.91         & 14.62         & 14.73           \\
Omni-Adjust~\cite{akimoto2022diverse}                     &  & 11.27         & 9.93          & 10.63           &  & 9.60          & 10.47         & 10.92           \\
ImmerseGAN~\cite{dastjerdi2022guided}  &  & 9.26         & 10.46          & 10.57           &  & 12.69          & 24.23         & 24.51           \\
PanoDiff (Ours)                  &  & \textbf{8.61} & \textbf{6.94} & \textbf{7.04}   &  & \textbf{7.72} & \textbf{6.96} & \textbf{7.18}   \\ \hline
\end{tabular}}
\end{table*}

After the sampling process, the decoder generates an image of shape $(w + 2w_p) \times h$, where $w_p$ is the extra width from the padded feature. To produce a standard panorama, the extra width is removed from the final image.

\section{Experiments}

\subsection{Implementation Details}
\subsubsection{Data Preparation}
We conducted experiments using real-world 360-degree panoramic image datasets SUN360~\cite{sun360} and Laval Indoor~\cite{lavalindoor}. SUN360 comprises both indoor and outdoor scenes, while Laval Indoor comprises solely indoor scenes. For SUN360, we randomly selected 2000/500 panoramas for training/testing, respectively. For Laval Indoor, we followed the approach in~\cite{hara2022spherical} and chose 289 images for testing. Notably, we \textbf{do not} train our model using the Laval Indoor dataset as there are already indoor scenes within our selected training data from the SUN360 dataset.

We used two input types in our experiments: a single input, which is a single NFoV image with a $90^\circ$ FoV placed at the center of the panorama, and paired input, which is a pair of NFoVs with relative rotation, allowing us to validate our method's capability to generate a panorama from multiple input images. We generated five pairs for each of our training/testing panoramas, resulting in 10,000/2,500 pairs of training/testing inputs on SUN360.

\subsubsection{Training}
We train our angle prediction network and latent diffusion model separately. 
The angle prediction network is trained with the strategies mentioned in ~\ref{sec:twostage}. Our Pano-denoiser and Pano-Encoder are trained for 7 epochs on pair inputs. For both single and pair input, we take in NFoV images of shape $256 \times 256$ and produce $1024 \times 512$ panoramas. During training, we generate input text prompts using BLIP~\cite{li2022blip}.

\subsubsection{Metrics}
We evaluate our method using two kinds of metrics. 

\noindent\textbf{Panorama Generation.} We use Fr\'echet Inception Distance (FID)~\cite{heusel2017gans, kynkaanniemi2022role} as our quantitative metric since FID can report the visual quality of generated panorama images to some extent. Besides, it has also been adopted by prior studies~\cite{hara2022spherical,akimoto2022diverse,wang2022stylelight}. 

\noindent\textbf{Rotation Estimation.} Following Cai \textit{et al.}~\cite{cai2021extreme}, we evaluate the geodesic error of the estimated rotation matrix (denoted by $\mathbf{\hat{R}}$) and the ground truth matrix (denoted by $\mathbf{R}$) using $\arccos(\frac{tr(\mathbf{R}^T \mathbf{\hat{R}})-1}{2})$. 

\begin{figure*}[t]
\begin{minipage}{0.95\textwidth}
  \centering
  \includegraphics[width=\textwidth]{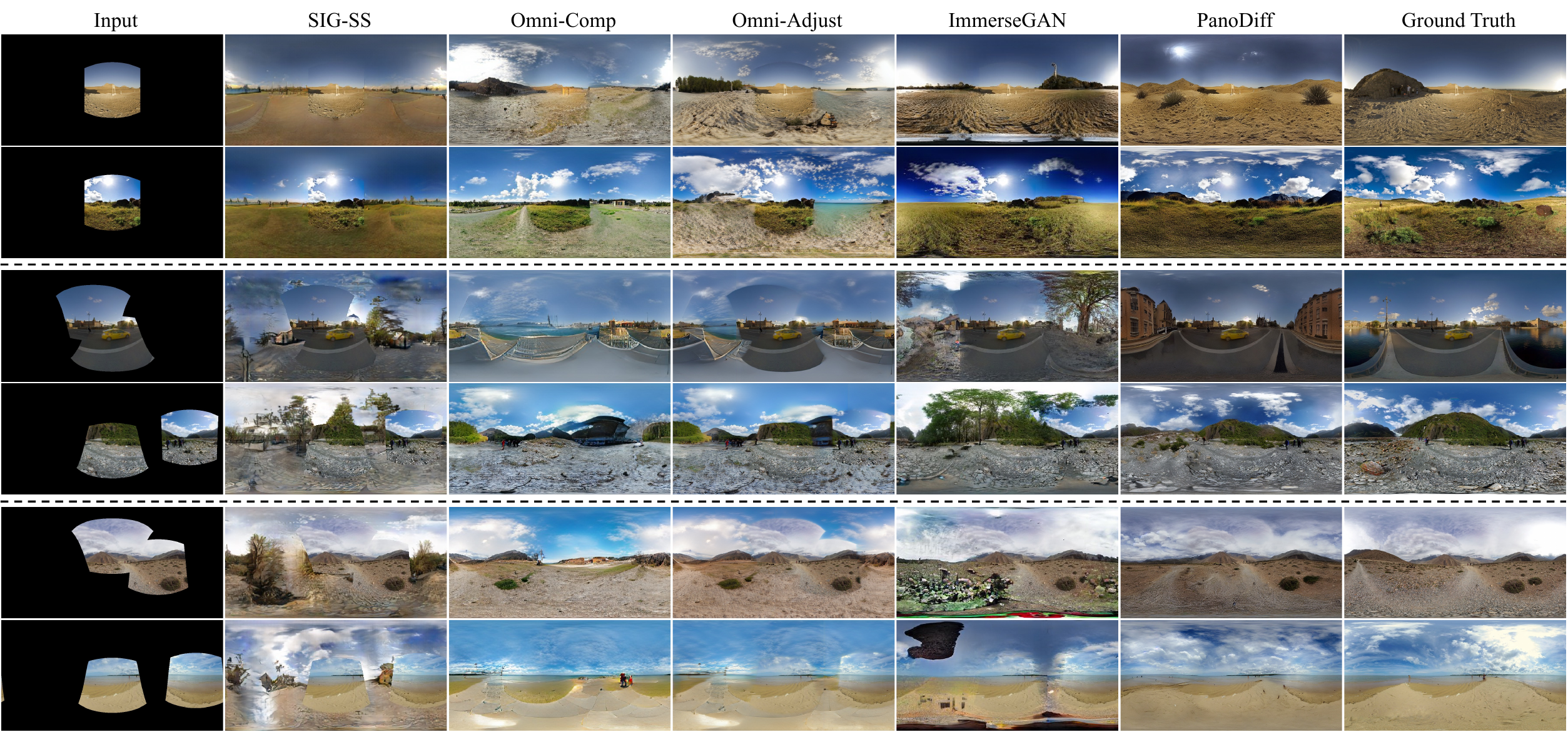}
  \caption{Out-painting results on SUN360 Dataset~\cite{sun360}. \textit{Omni-Comp} and \textit{Omni-Adjust} denote the CompletionNet and AdjustmentNet outputs of Omni-Dreamer~\cite{akimoto2022diverse}, respectively. }
  \label{fig:exp_comparison}
\end{minipage}
\begin{minipage}{0.95\textwidth}
  \centering
  \includegraphics[width=\textwidth]{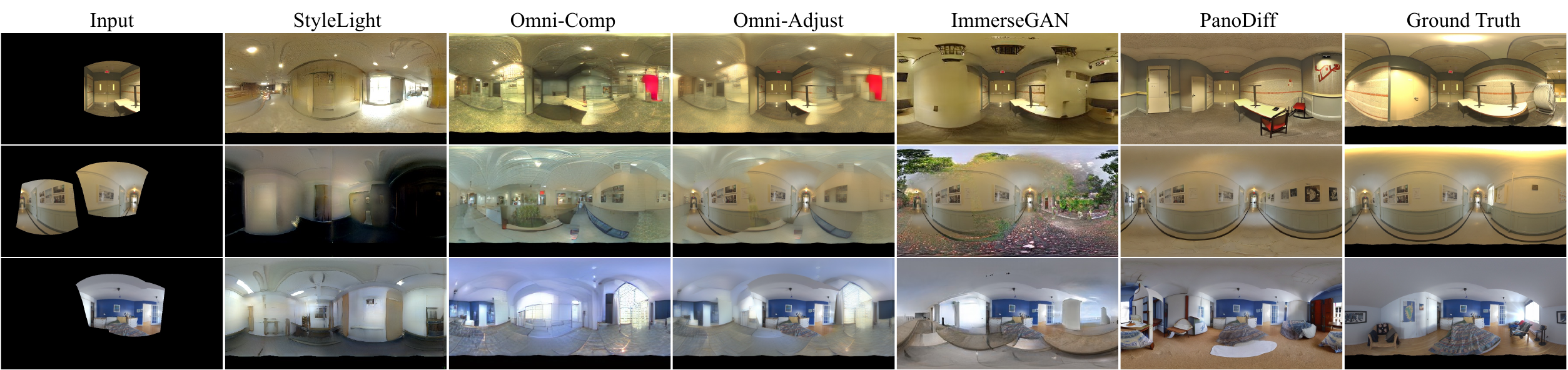}
  \caption{Out-painting results on Laval Indoor Dataset~\cite{lavalindoor}. We evaluate the models of \textit{StyleLight}, \textit{Omni-Comp}, and \textit{Omni-Adjust}, by utilizing their officially released models trained on Laval Indoor Dataset. The results obtained from \textit{ImmerseGAN} were graciously provided by the authors who ran their model for our evaluation.
  }
  \label{fig:exp_comp_laval}
\end{minipage}
\end{figure*}

\subsubsection{Baselines}
We compared our method to three previous SOTA methods. Omni-Dreamer~\cite{akimoto2022diverse} is trained on SUN360 with 47938 images. For the Laval dataset, Omni-Dreamer is finetuned on the model trained on SUN360 and with 1837 training samples. SIG-SS~\cite{hara2022spherical} is only trained on SUN360 with 50000 training images. Stylelight~\cite{wang2022stylelight} is trained on the Laval dataset, the same training set as used in Omni-Dreamer. We inferred these models with their officially released trained models on our test split. Since their training split is unknown to us, images in our test set could be in their training set. We also reached out to the authors of ImmerseGAN~\cite{dastjerdi2022guided} and acquired their model's results on our test set.


\begin{table}[!t]
\caption{Evaluation of relative pose estimation on SUN360~\cite{sun360} and Laval Indoor~\cite{lavalindoor}. We present the Average geodesic error Avg($^\circ$$\downarrow$) and the percentage of pairs with errors under 10 degrees 10$^\circ$(\%$\uparrow$). }
\label{tab:rota}
\resizebox{\linewidth}{!}{
\begin{tabular}{ccccp{0.02cm}cc}
\hline
                               &            &   \multicolumn{2}{c}{SUN360~\cite{sun360}}      &  & \multicolumn{2}{c}{Laval~\cite{lavalindoor}}       \\ \cline{3-4} \cline{6-7} 
Pair Type                      & Method     &   Avg($^\circ$$\downarrow$)            & 10$^\circ$(\%$\uparrow$)           &  & Avg($^\circ$$\downarrow$)            & 10$^\circ$(\%$\uparrow$)           \\ \hline
\multirow{2}{*}{Overlap}       & 1-stage &   7.19           & 90.27          &  & 4.21           & 96.74          \\
                               & 2-stage       &   \textbf{3.58}  & \textbf{95.61} &  & \textbf{3.66}  & \textbf{96.88} \\ \hline
\multirow{2}{*}{\makecell{Wide Baseline}} & 1-stage &  41.52          & 40.64          &  & 32.31          & 62.34          \\
                               & 2-stage       &   \textbf{27.12} & \textbf{49.77} &  & \textbf{29.46} & \textbf{62.62} \\ \hline
\multirow{2}{*}{All}           & 1-stage &   24.29          & 65.54          &  & 18.00          & 79.86          \\
                               & 2-stage        & \textbf{15.31} & \textbf{72.77} &  & \textbf{16.32} & \textbf{80.07} \\ \hline
\end{tabular}}
\end{table}

\subsection{Quantitative Evaluation}
We examine the performance of our approach from two perspectives, namely the accuracy of rotation estimation and the panorama generation quality. 

\noindent\textbf{Rotation Estimation.} We evaluate the performance of relative rotation estimation on SUN360~\cite{sun360} and Laval Indoor~\cite{lavalindoor} datasets. For clarity, we separate the input pairs into two categories based on whether they belong to the 'overlap' or 'wide baseline'. We compare two relative rotation estimation models: our proposed two-stage model, and a single-stage model which is designed the same as ~\cite{cai2021extreme}.
Table. ~\ref{tab:rota} reports the evaluation results, which demonstrate that the two-stage relative rotation estimation method significantly outperforms the single-stage approach in terms of average relative pose estimation error for both `overlap' and `wide baseline'. It is noted that both methods trained only on the SUN360 dataset.

\noindent\textbf{Panorama Generation. }
The primary results are summarized in Table. ~\ref{tab:methodcompare}. As demonstrated in the table, our method attains the most favorable FID metrics with all three input types on both datasets. The terms \textit{GT rots} and \textit{Pred rots} denote the utilization of ground truth relative angles and predicted angles from our network $\mathcal{R}_\theta$, respectively. Despite the imperfect nature of the FID metric~\cite{parmar2021cleanfid, kynkaanniemi2022role}, the significant margin of our method's superiority substantiates its overall effectiveness. It is noteworthy that our model is NOT trained on the Laval Indoor dataset, yet it surpasses the performance of previous methods that were specifically trained on this dataset.

\subsection{Qualitative Comparison}

The visual results are shown in Figure ~\ref{fig:exp_comparison}. To illustrate the generation quality of our method, we compare our method to previous works under three kinds of inputs: single NFoV input(top two rows), the paired NFoV input with \textit{GT rots}(middle two rows), and the paired NFoV input with \textit{Pred rots}(bottom two rows). 
As can be observed from the results, both SIG-SS and Omni-Adjust exhibit inconsistencies in generating boundaries between the input patches and the out-painting content.  While the generated images of Omni-Comp show a marginal improvement in boundary issues, they compromise the quality and realism of the generated content to achieve smooth output boundaries.  Omni-Adjust produces relatively more realistic content, but it occasionally out-paints the wrong content. For instance, in the second row, Omni-Adjust wrongly out-paints a `beach' while the input region indicates a grassland scene. ImmerseGAN excels at generating satisfactory results with a single FoV input, but struggles with paired FoV input due to the lack of specific training for handling multiple FoVs. PanoDiff outperforms previous approaches in content consistency, realism, and texture continuity, maintaining high-quality and realistic content generation.

\begin{figure*}[t]
    \centering
    \includegraphics[width=0.95\linewidth]{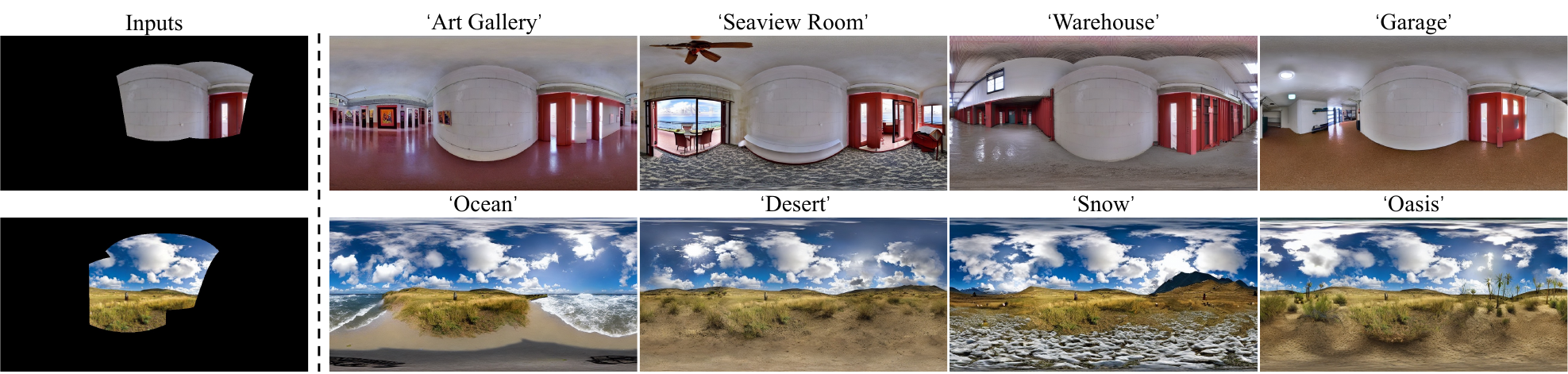}
    \caption{Control panorama generation with text prompts. This demonstrates the capability of our method to generate diverse high-quality results with various text prompts, while still able to maintain the geometric characteristics of panoramas.}
    \label{fig:edit}
\end{figure*} 

We proceed to look at the generalizing performance of our approach on Laval Indoor Dataset\cite{lavalindoor} and present the results in Figure~\ref{fig:exp_comp_laval} with both single and paired inputs. Note that our model was NOT trained on the Laval Indoor dataset but solely on SUN360. Nonetheless, our method still achieves high-quality panorama generation with consistent visual properties, \emph{e.g.}, lighting, temperature, and geometrical consistency.

\vspace{-0.3cm}
\subsection{Ablation Study}
We validate the efficacy of key components in our approach, \emph{i.e.}, the denoising strategies, and the circular padding strategy.  
\begin{figure}[h]
    \centering
    \includegraphics[width=\linewidth]{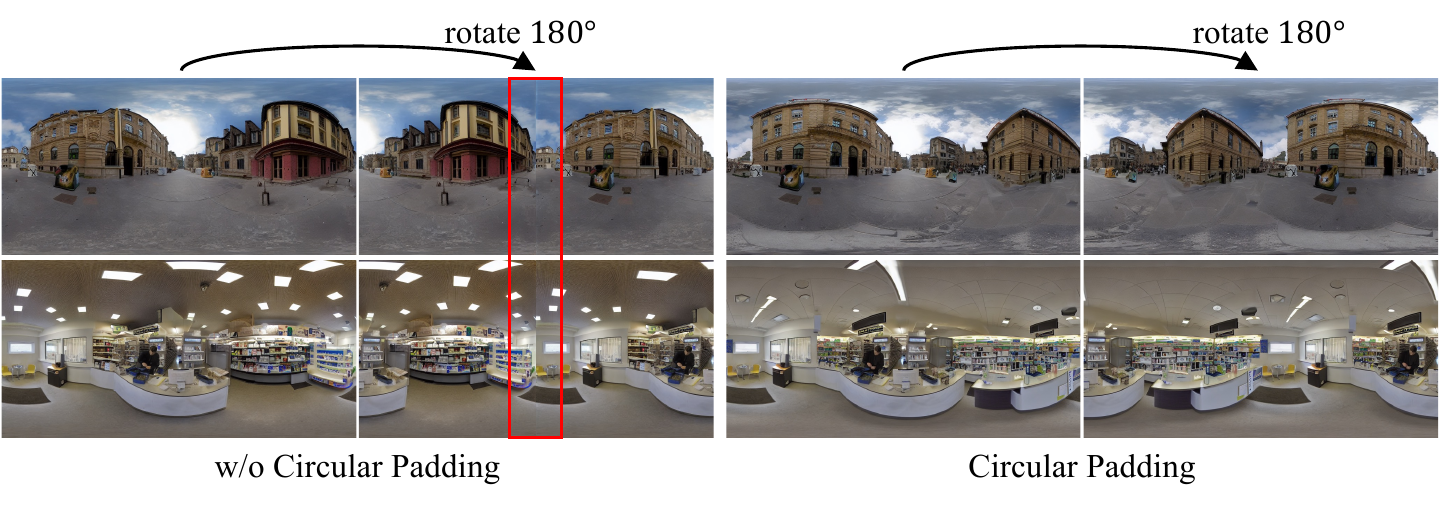}
    \caption{Circular padding solves the problem of discontinuity between the left and right sides in panorama generation.}
    \label{fig:ablation_padding}
\end{figure}


\noindent\textbf{Denoising Strategies in 360-Degree.} We conduct quantitative comparisons in Table. ~\ref{tab:abl_strategy} regarding the usage of our rotation equivariance loss, and rotating schedule. These denoising strategies assist our model to understand the geometric characteristics of panoramas in latent space. As could be seen from the table, both our strategies contribute significantly to the quality of our panorama generation. 
\begin{table}[h]
\caption{FID$\downarrow$ results of PanoDiff with different strategies. `Equi.' denotes the usage of our rotation equivariance loss in Sec.~\ref{sec:equi}, and `Schedule' denotes our rotating denoising schedule in Sec. ~\ref{sec:schedule}.}
\label{tab:abl_strategy}
\begin{tabular}{c|cc|c}
\hline
Strategy   &  Equi.  & Schedule  & FID $\downarrow$ \\
\hline
\multirow{3}{*}{Choices}& - & - & 7.88 \\
& $\checkmark$ & - & 6.73 \\
& $\checkmark$ & $\checkmark$ & \textbf{6.56} \\
\hline
\end{tabular}
\end{table}

\noindent\textbf{Circular Padding For Continuity.} When decoding the latent feature $z_0$, the inherited border effect caused by the domain gap in latent space and image space frequently occurs. To alleviate this issue, we implement a circular padding strategy as described in ~\ref{sec:cirpad}.  
In Figure~\ref{fig:ablation_padding}, we validate the necessity of circular padding with visual result comparisons by rotating $180^\circ$ horizontally from the output image. As observed, circular padding seamlessly eliminates discontinuity between both ends.

\subsection{Applications}
\noindent\textbf{Text Editing. }
We extend the editing capability of our method. As depicted in Figure~\ref{fig:edit}, our method not only inherits the powerful text-to-image generation ability from Stable Diffusion but also preserves the sound geometric properties of panoramas.

\begin{figure}[ht]
    \centering
    \includegraphics[width=\linewidth]{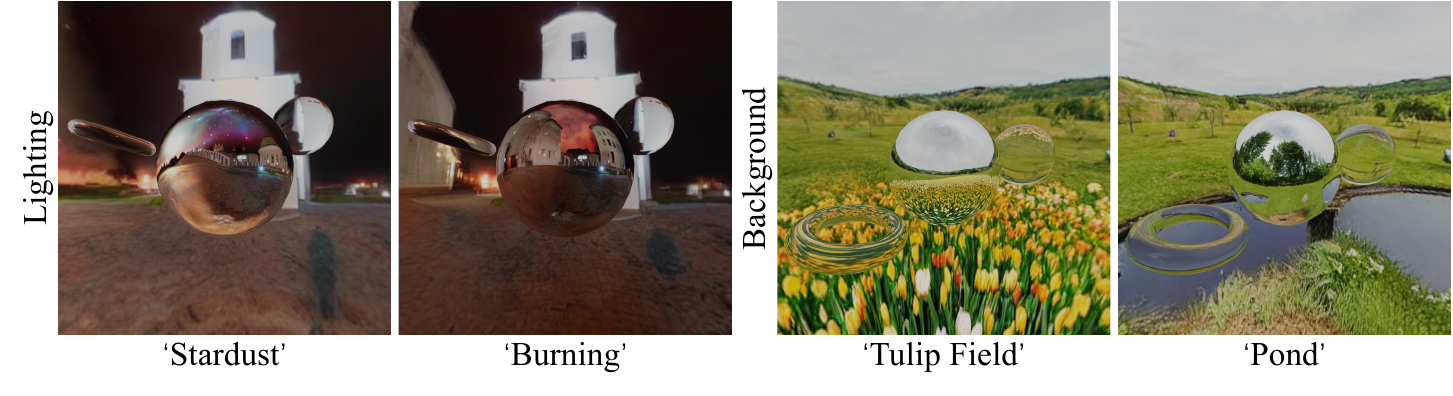}
    \caption{Examples of panoramic images used as environment textures. }
    \label{fig:blender}
\end{figure}

\noindent\textbf{Environment Texture. }
Panoramic images can serve as environment textures in 3DCG software, which can provide background lighting for 3D assets. Figure~\ref{fig:blender} shows some examples of our generated panoramas as environmental textures. As can be found, our method produces diverse panoramas that not only serve as plausible rendering backgrounds (first two) but also provide environmental lighting (last two).

\noindent\textbf{Multiple NFoV As Input. }
Our framework includes a two-stage relative pose estimation module, allowing it to handle multiple NFoV images (e.g., >2) of the same scene as input. 
As shown in Figure~\ref{fig:teaser}, our pipeline can robustly generate high-quality panoramic images using multiple NFoV images. Please refer to the supplementary for more generated 360-degree panoramas.
\section{Conclusion}
In this paper, we present PanoDiff, a novel framework that generates 360$^\circ$ panoramas from one or more NFoV inputs. The pipeline consists of two main modules, namely rotation estimation and panorama generation. Our two-stage rotation estimation network first classifies the input image pairs into overlap and wide baseline scenarios and then performs precise angle prediction. In panorama generation, we use incomplete partial panoramas along with text prompts as signals to generate diverse panoramas. 
We hope that our work inspires further research in panorama generation for advanced applications, including style control, direct HDRI panorama generation, and related areas.

\noindent \textbf{Acknowledgement.} This work was supported by the Fundamental Research Funds for the Central Universities, STCSM under Grant 22DZ2229005, 111 project BP0719010.

\clearpage
\newpage


\bibliographystyle{ACM-Reference-Format}
\balance
\bibliography{sample-base}

\newpage

\end{document}